\documentclass[sigconf,nonacm]{aamas}  
\acmBooktitle{}
\acmConference[]{}{}{}{}
\acmConference[]{}{}{}
\acmBooktitle{}
\acmDOI{}
\acmISBN{}

\usepackage{balance} 

\usepackage{booktabs}
\usepackage{adjustbox}
\usepackage[table]{xcolor}
\usepackage{mathtools}

\definecolor{vlmbest}{HTML}{FFF1BF}   
\definecolor{basebest}{HTML}{CDE7FF}  
\newcommand{\wvlm}[1]{\cellcolor{vlmbest}{#1}}
\newcommand{\wbase}[1]{\cellcolor{basebest}{#1}}
\usepackage{amsmath} 
\usepackage{enumitem} 
\newcommand{\HP}{\mathrm{HP}}
\newcommand{\DMG}{\mathrm{DMG}}
\DeclareMathOperator{\sign}{sign}
\newcommand{\reg}[1]{\textsc{#1}}   

\setcopyright{ifaamas}
\acmConference[AAMAS '26]{Proc.\@ of the 25th International Conference
on Autonomous Agents and Multiagent Systems (AAMAS 2026)}{May 25 -- 29, 2026}
{Paphos, Cyprus}{C.~Amato, L.~Dennis, V.~Mascardi, J.~Thangarajah (eds.)}
\copyrightyear{2026}
\acmYear{2026}
\acmDOI{}
\acmPrice{}
\acmISBN{}

\usepackage{colortbl}
\usepackage{tabularx,multirow}
\usepackage{siunitx}  
\usepackage[most]{tcolorbox} 
\usepackage{fvextra}

\definecolor{lightgreen}{RGB}{200, 255, 200}
\definecolor{lightpurple}{RGB}{220, 200, 255}

\definecolor{lightblue}{RGB}{200, 220, 255}   
\definecolor{lightred}{RGB}{255, 200, 200}    

\definecolor{beige}{RGB}{245, 245, 220}
\definecolor{darkred}{RGB}{200, 0, 0}
\arrayrulecolor{black}  

\acmSubmissionID{1297}

\title[AAMAS-2026 Formatting Instructions]{StarBench: A Turn-Based RPG Benchmark for Agentic Multimodal Decision-Making and Information Seeking}

\author{Haoran Zhang}
\authornote{Equal contribution}
\affiliation{
  \institution{University of Michigan}
  \city{Ann Arbor}
  \country{United States}}
\email{haoranwh@umich.edu}

\author{Chenhao Zhu}
\authornotemark[1]
\affiliation{
  \institution{Stanford University}
  \city{Stanford}
  \country{United States}}
\email{chenhzhu@stanford.edu}

\author{Sicong Guo}
\affiliation{
  \institution{University of Michigan}
  \city{Ann Arbor}
  \country{United States}}
\email{stevengu@umich.edu}

\author{Hanzhe Guo}
\affiliation{
  \institution{University of Michigan}
  \city{Ann Arbor}
  \country{United States}}
\email{hanzheg@umich.edu}

\author{Haiming Li}
\affiliation{
  \institution{University of Michigan}
  \city{Ann Arbor}
  \country{United States}}
\email{haiming@umich.edu}

\author{Donglin Yu}
\affiliation{
  \institution{University of Illinois Urbana–Champaign}
  \city{Urbana–Champaign}
  \country{United States}}
\email{donglin5@illinois.edu}

\begin{abstract}
Human players do more than press buttons: they ground what they \emph{see} on screen into precise keyboard–mouse actions and, when stuck, they \emph{seek information} before trying again. We ask whether current vision–language models (VLMs) can do the same. Despite encouraging results under simplified control or tool scaffolds, \emph{human-like play in a real client}—mapping raw screenshots to temporally coherent low-level actions while deciding when to ask for guidance—remains an open challenge. We introduce \textbf{StarBench}, a turn-based RPG benchmark derived from \emph{Honkai: Star Rail} that targets these two human-like competencies: \emph{multimodal decision-making from pixels to actions} and \emph{agentic information seeking}. StarBench standardizes evaluation across eight combat tasks and two regimes with shared tasks and metrics: (i) \emph{direct control}, where agents receive only screenshots and must emit low-level primitives (click and keypress) with no semantic hints; and (ii) \emph{tool-assisted control}, where higher-level intents can be mapped to primitives by detectors and OCR outputs provide optional textualized observations to ease UI grounding. To mirror human practice, StarBench also includes an \emph{ask-or-act} diagnostic that measures \emph{whether and when} agents choose to request brief guidance before proceeding, and how that choice affects subsequent performance. We report reference baselines for contemporary VLMs and a human reference. Results expose sizable gaps in perception-to-control fidelity in the direct regime, while showing that judicious information seeking correlates with improved success—establishing StarBench as a reproducible yardstick for agentic information seeking and multimodal decision-making in real-client play.
\end{abstract}


\newcommand{\BibTeX}{\rm B\kern-.05em{\sc i\kern-.025em b}\kern-.08em\TeX}

\begin{document}


\pagestyle{fancy}
\fancyhead{}


\maketitle 

\begin{table*}[t]
\centering
\caption{Comparison of game-centric evaluation settings. \textbf{StarBench} uniquely pairs a \emph{direct screenshot-to-action} track with a matched \emph{tool-assisted} track under identical tasks/metrics, and adds an \emph{ask-or-act} decision diagnostic.}
\label{tab:benchmark-compare}
\resizebox{\linewidth}{!}{
\begin{tabular}{lcccccc}
\toprule
& \multicolumn{2}{c}{\textbf{Environment}} & \textbf{Control Interface} & \textbf{Scaffolds} & \multicolumn{2}{c}{\textbf{Tracks \& Diagnostics}} \\
\cmidrule(lr){2-3} \cmidrule(lr){5-5} \cmidrule(lr){6-7}
\textbf{Benchmark / Setting} &
\textbf{Real Client} &
\textbf{Observation} &
\textbf{Action Interface} &
\textbf{Tool Assist} &
\textbf{Direct Ctrl. Track} &
\textbf{Ask-or-Act} \\
\midrule
Atari ALE \citep{mnih2013playing}           & Emulator         & Pixels        & Low     & No      &  Yes & No \\
StarCraft II (RL) \citep{liu2022efficient}  & Engine API       & Pixels+Text        & High           & No      & No                         & No \\
MineDojo / Voyager \citep{fan2022minedojo,wang2023voyager}
                                            & Real Client + Engine API   & Pixels+Text   & High       & Yes   & No                         & No \\
Pok\'eLLMon \citep{hu2024pokellmon}         & Engine API       & Text    & High    & No     & No                         & No \\
BALROG \citep{paglieri2023balrog}           & Emulator    & Pixels+Text   & High          & Yes   & No                         & No \\
CRADLE (RDR2) \citep{tan2024cradle}         & Real Client      & Pixels+Text   & High       & Yes   & No                         & No \\
MiniHack \citep{samvelyan2021minihack} & Engine API   & Text           & High   & No     & No                         & No \\
NetHack (NLE) \citep{kuttler2020nethack} & Engine API & Text           & High   & No     & No                         & No \\
\textbf{StarBench (ours)}                   & \textbf{Real Client} & \textbf{Pixels (\& optional Text)} & \textbf{Low/High} & \textbf{Optional} & \textbf{Yes} & \textbf{Yes} \\
\bottomrule
\end{tabular}
}
\raggedright\footnotesize
\textbf{Real Client}: whether actions are sent to a real game window via OS events (vs.\ emulator/engine APIs).
\textbf{Observation}: what the agent perceives (raw screenshots “Pixels”; Engine/API returned state information/OCR summaries “Text”).
\textbf{Action Interface}: Low-level OS primitives like \texttt{click/keys} vs.\ High-level scripted/engine APIs or macros).
\textbf{Tool Assist}: availability of helpers that simplify perception or action selection (e.g., textualized UI summaries and detector/OCR JSON).
\textbf{Direct Ctrl.\ Track}: presence of a pure screenshot$\rightarrow$primitive setting with \emph{no} semantic aids.
\textbf{Ask-or-Act}: whether the setting includes a controlled decision point to \emph{ask} for concise guidance or \emph{proceed} without it.\\
\end{table*}
\section{Introduction}
With the emergence of large language models (LLMs) such as ChatGPT \citep{ouyang2022training}, research on intelligent agents has expanded from language-centric tasks (e.g., code generation \citep{jiang2024survey} and conversational systems \citep{friedman2023leveraging}) to interaction with dynamic environments \citep{wang2024large}. This shift raises a direct question: \emph{can current VLMs play games like humans do}---perceiving raw screens, deciding, and issuing precise keyboard-mouse actions---\emph{without} relying on bespoke macros or scripted APIs?

Games offer a compelling testbed for this question: they are structured yet diverse, repeatable yet challenging. Prior work spans traditional board games and modern role-playing games (RPGs), from \emph{Pokémon} \citep{hu2024pokellmon} to classic titles like \emph{Go} and \emph{Chess} \citep{feng2023chessgpt,wang2025are}. Early evaluations often targeted unimodal agents—purely visual \citep{mnih2013playing} or purely textual \citep{cote2019}. More recent efforts incorporate multimodal inputs, but many rely on simplified simulators, scripted APIs, or macros, which blur whether a model can truly ground pixels into low-level control.

Answering the human-like play question requires a benchmark that \emph{separates} two factors yet evaluates them under the \emph{same} tasks and metrics. First, \textbf{direct screenshot-to-action control}: the agent sees only the screenshot and must output low-level primitives (click and keypress) at precise coordinates. Second, \textbf{information access}: humans routinely decide whether to look up information after setbacks; thus, beyond raw control, a benchmark should measure \emph{when} an agent chooses to seek information versus proceeding to act.

\emph{Honkai: Star Rail} (HSR) is a turn-based RPG whose squad-based combats naturally intertwine visual cues (enemy states, turn order) and textual descriptions (skills, mechanics, character specs). Its complexity and evolving UI patterns make it a strong platform for studying perception-to-control links in realistic clients. Importantly, HSR is a \emph{continually updated} live-service game: periodic balance changes, new enemies, and UI adjustments reduce the utility of static prior knowledge embedded in pretrained models. This dynamic setting naturally pressures agents to decide when to \emph{seek task-relevant information} (e.g., clarifying mechanics or target priorities) versus proceeding without it.

We introduce \textbf{StarBench}, a benchmark derived from HSR that evaluates multimodal agents in a \emph{real game client} across eight combat tasks. StarBench standardizes both \textbf{direct control} (screenshot $\rightarrow$ keyboard-mouse) and \textbf{tool-assisted control} (computer-operation primitives with textualized observations) under identical task definitions, success criteria, and configurations for reproducibility. To analyze the decision of seeking information, StarBench further includes an \textbf{ask-or-act} diagnostic experiment: before each battle, agents may either \textsc{ask} a targeted question to obtain bounded textual guidance or \textsc{continue} without assistance; we measure ask rate, timing, benefit, and downstream performance. Our contribution could be summarized as following:
\begin{enumerate}
    \item \textbf{Benchmark specification.} We present \textit{StarBench}, a single-title yet multi-task benchmark in HSR with a precise environment interface, eight tasks spanning distinct combat competencies, and standardized metrics and configurations for reproducibility.
    \item \textbf{Two evaluation regimes.} We formalize protocols for both \emph{direct screenshot-to-action control} and \emph{tool-assisted control} (computer-operation primitives with textualized observations).
    \item \textbf{Ask-or-act decision diagnostic.} We provide a method-agnostic diagnostic that quantifies whether and \emph{when} agents decide to seek information (ask) versus proceed (act), and how these choices affect subsequent gameplay.
    \item \textbf{Reference baselines.} We report reference results for contemporary VLMs and a human reference to establish starting points rather than state-of-the-art claims.
\end{enumerate}


\section{Related Work}

\subsection{Vision-Language Models}

Vision-language models (VLMs) have emerged as a powerful paradigm for integrating visual and textual modalities, allowing for a variety of applications, including image captioning \citep{vinyals2015show,xu2015show}, visual question answering \citep{antol2015vqa}, and cross-modal retrieval \citep{li2023cross}. Early approaches, such as CLIP \citep{radford2021learning}, leverage contrastive learning to align visual and language representations, thereby achieving robust zero-shot performance in various classification tasks. More recent work extends these methods to generate textual output conditioned on visual inputs. For instance, models such as GPT-4V \citep{openai2023gpt4}, Gemini \citep{gemini2024}, and Claude-3.5-Sonnet \citep{anthropic2024claude35} employ cross-attention mechanisms \citep{lin2021cat} to fuse features from vision transformers \citep{dosovitskiy2020image,vaswani2017attention} with large language models. This architectural integration not only facilitates the generation of detailed image descriptions, but also enhances contextual understanding in downstream tasks. 
\begin{figure*}[t]
    \centering
    \includegraphics[width=\linewidth]{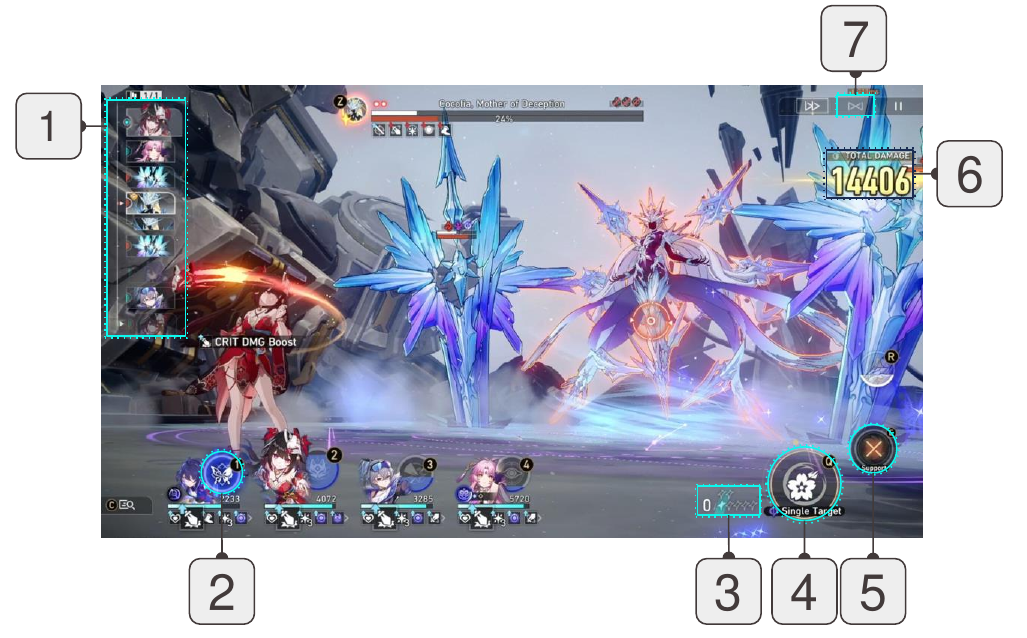}
    \caption{Annotated HSR combat UI. Key affordances include: (1) the \emph{action order} track governing future turns; (2) \emph{Ultimate} indicators (glow = ready); (3) team \emph{skill points}; (4) \emph{Basic Attack} and (5) \emph{Skill} buttons with availability reflecting (3); (6) damage and status text; (7) the auto-battle toggle. Top-center, the boss frame shows a white \emph{toughness} bar above a red \emph{HP} bar. In the example, the team has \texttt{0} skill points, disabling (5) while (4) remains usable; the glow at (2) indicates Seele’s Ultimate can be cast, potentially off-turn.}
    \label{fig:ui_components}
     \Description{Screenshot of the HSR combat UI with annotations for key interface elements.}
\end{figure*}

\subsection{Games as Benchmarks}
Since Arthur Samuel's seminal work in 1959 on checkers-playing programs \citep{samuel1959studies}, games have served as fundamental benchmarks for evaluating artificial intelligence systems \citep{hu2024survey,sweetser2024llmgames}. The evolution of game-based benchmarks mirrors advances 
in AI capabilities, from rule-based board games like \emph{Chess} that challenged early symbolic systems \citep{shannon1950programming}, to real-time video games like \emph{Atari} games and \emph{StarCraft II} that tested reinforcement learning approaches \citep{mnih2013playing, liu2022efficient}. Modern interactive environments such as \emph{Minecraft} \citep{johnson2016malmo,fan2022minedojo,wang2023voyager} and \emph{RDR2} \citep{tan2024cradle} now push the boundaries of multimodal understanding and long-horizon planning. These gaming environments offer three key advantages as AI benchmarks: (1) structured rule systems that enable quantitative evaluation, (2) complexity that mirrors real-world challenges, and (3) multimodal interfaces that require human-like perception and reasoning. While these settings span a broad spectrum of observations and action interfaces, they often conflate perception, control, and assistance. Table~\ref{tab:benchmark-compare} contrasts representative evaluations and highlights how \textbf{StarBench} different from existing game benchmarks.

\subsection{VLMs as Game Agents}
Traditional game agents often assume privileged interfaces—direct APIs to game state \citep{wang2023voyager,hu2024pokellmon} or unimodal inputs \citep{xu2023exploring}—that bypass many challenges human players face. Humans, by contrast, must ground actions in raw pixels, resolve UI affordances, and combine on-screen perception with textual knowledge, all under partial observability. Recent VLM-based agents push toward this human-like setting by reasoning over screenshots and language, but common evaluations still blur key difficulties. Some operate in simplified simulators or expose high-level action APIs that hide low-level control; others lean on large, static knowledge stores or extensive demonstrations \citep{chen2024canvlm,paglieri2023balrog}. This makes it difficult to disentangle (i) perception-to-control fidelity from (ii) the benefits of tool scaffolding or textual hints. Our benchmark makes this concrete by standardizing two regimes under identical tasks and metrics: \emph{direct control}, where the agent sees only screenshots and must emit low-level keyboard-mouse primitives (no semantic aids), and \emph{tool-assisted control}, where the agent may get help from tool for easier perception and manipulation while playing the game (like previous benchmarks). We also include an \emph{ask-or-act} diagnostic that measures whether and when an agent seeks concise guidance before proceeding. This design isolates core capabilities for human-like play without relying on privileged game APIs.

\section{Background}

\subsection{Problem Formalization}
We model \emph{Honkai: Star Rail} (HSR) combat as a partially observable Markov decision process (POMDP) \(\mathcal{M}=(\mathcal{S},\mathcal{A},\mathcal{T},\mathcal{R},\mathcal{O})\). The latent state \(\mathcal{S}\) comprises complete combat information—ally and enemy health, energy, toughness, speed, buffs/debuffs, and internal cooldowns—while the observation space \(\mathcal{O}\) exposes only a subset of this information through the on-screen user interface (UI), making perception inherently partial and noisy (see Figure~\ref{fig:ui_components} for an annotated overview of the combat screen.). The action space \(\mathcal{A}\) contains all feasible player commands, and the transition kernel \(\mathcal{T}(s_{t+1}\!\mid s_t,a_t)\) captures both deterministic rules (e.g., turn scheduling) and stochastic effects (e.g., critical hits, status proc chances). The reward function \(\mathcal{R}(s_t,a_t)\) is task-dependent (e.g., boss defeat under step limits, score accumulation under action-value budgets) and is used to evaluate outcomes rather than to define additional supervision. 

\subsection{HSR Combat Mechanics}
\begin{figure*}
    \centering
    \includegraphics[width=1\linewidth]{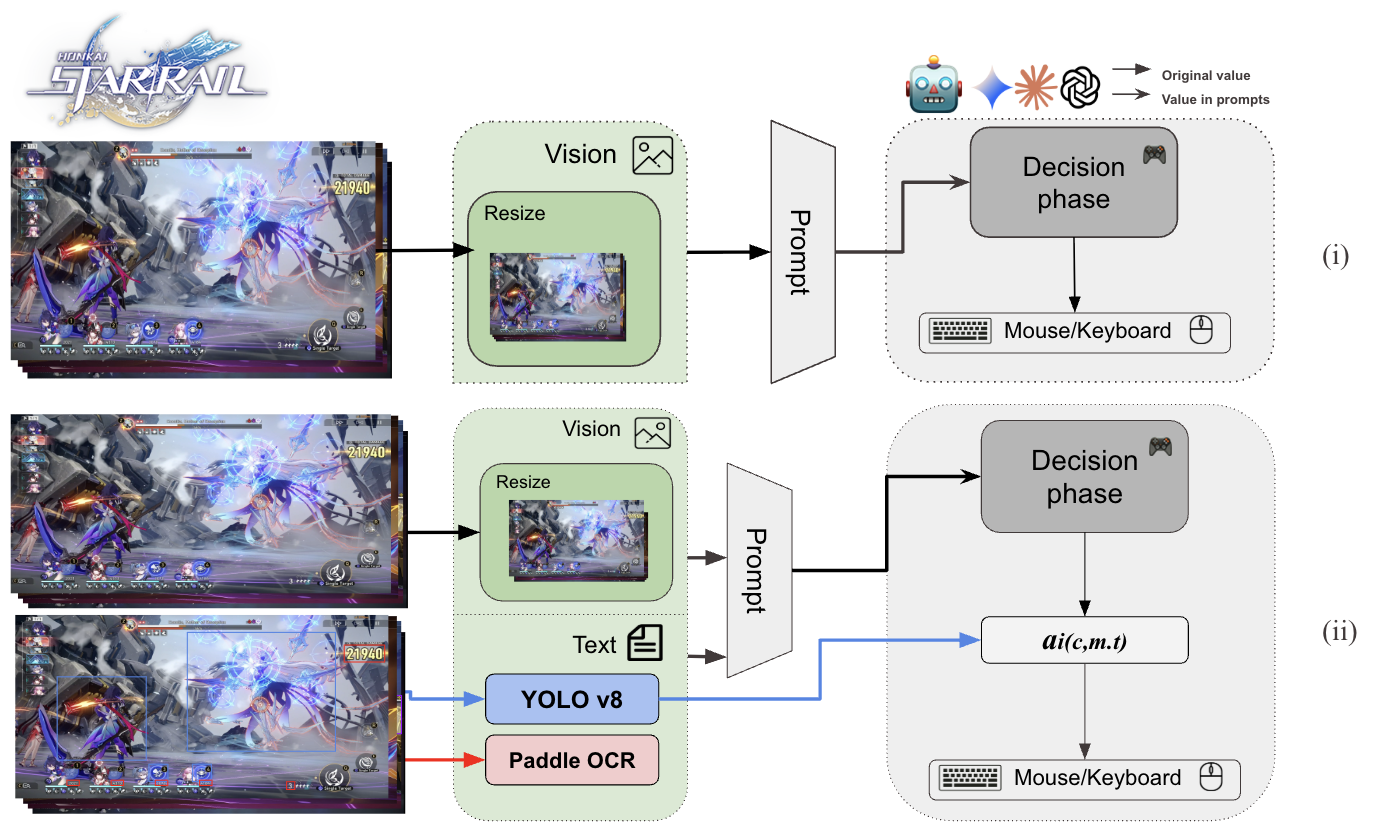}
  \caption{Interaction Protocol: (i) DC (direct control) and (ii) TA (tool-assisted). Inputs are resized to fit each VLM’s native resolution, and outputs are resized back to the original scale for evaluation.}
    \label{fig:pipeline}
\end{figure*}
HSR features squad-based, turn-driven combat in which a player controls up to four characters against one or more enemies. Turn order is governed by speed-related statistics and evolves over time due to buffs, debuffs, and “Weakness Break” effects. On a character’s turn, the player typically chooses among a \emph{Basic Attack}, a \emph{Skill}, or (when available) an \emph{Ultimate}. Skills consume a shared team resource—\emph{skill points}—while Basic Attacks replenish it, creating a tight \emph{economy of actions} that trades short-term power for future flexibility. Ultimates are gated by a separate \emph{energy} meter and may be cast \emph{off-turn} as an interrupt when fully charged, adding an additional layer of timing and coordination.

Enemy resilience is mediated by a \emph{toughness} bar and \emph{elemental weaknesses}. When attacks of an appropriate element reduce an enemy’s toughness to zero, the enemy enters \emph{Weakness Break}, experiencing delayed turns and reduced resistances. Managing break windows—while maintaining team survivability—requires fusing visual cues (e.g., turn order track, HP/energy bars, toughness and weakness icons, ability readiness) with textual information (e.g., skill and status descriptions). Because only a subset of internal variables is directly visible, agents must infer latent quantities from these on-screen signals.

Throughout this work we consider a canonical Quantum-aligned team to keep evaluation controlled and reproducible. In our configuration, a primary damage dealer (e.g., \emph{Seele}) can chain eliminations via an extra-turn mechanic; a support (e.g., \emph{Sparkle}) advances ally turns and modulates the team’s damage and skill-point economy; a debuffer (e.g., \emph{Silver Wolf}) implants weaknesses and reduces defenses; and a defender (e.g., \emph{Fu Xuan}) mitigates incoming damage and provides team-wide sustain. This composition illustrates the interplay among resource management (skill points and energy), turn manipulation, and weakness exploitation that underlies strategic decision making in HSR.


\section{StarBench}

\subsection{Interaction Protocol}
\textit{StarBench} evaluates real-client play under two matched regimes that share the same game client, tasks, and metrics, and differ only in how actions are expressed and how much UI grounding assistance is provided.(see Figure.~\ref{fig:pipeline})

\subsubsection{Direct Control (DC)}
The agent receives the raw screenshot at a fixed resolution and UI scale (\(1920\times1080\)) and must act purely by spatial pointing. At each decision step it must (i) return an \emph{explicit click location} as \emph{pixel coordinates} \((x,y)\) with \(x\in[0,1919],\,y\in[0,1079]\) for select target and (ii) a keypress for confirming the action.
The environment then issues OS primitives via \texttt{pyautogui} library \cite{pyautogui}. Out-of-bounds values are clipped to the valid range; empty or non-numeric outputs are treated as no-ops. No YOLO/OCR summaries or other semantic aids are provided—performance reflects pixel-level UI parsing and localization.

\subsubsection{Tool-Assisted (TA)}
\begin{table}[htbp]
\label{example_action_space}
\centering
\begingroup
\setlength{\tabcolsep}{10pt}    
\renewcommand{\arraystretch}{1.35} 

\small 

\begin{tabular}{l|p{0.64\linewidth}} 
\hline
\textbf{Component} & \textbf{Description} \\
\hline
Character (\(c\)) & 0: Seele; 1: Sparkle; 2: Silver Wolf; 3: Fu Xuan \\
\hline
Move (\(m\)) & 0: Basic Attack; 1: Skill; 2: Release Ultimate;   
3: Hold Ultimate \\
\hline
Target (\(t\)) & 0--3: Teammates; 4--8: Enemies; 9: Select all \\
\hline
\end{tabular}
\caption{Action space in StarBench under TA Mode. Teammates organized based on characters position in team (e.g., Seele is the left most character so it is at position 0). Hold Ultimate represents not release the Ultimate in off-turn. }
\label{tab:action_space}
\endgroup
\end{table}
The same screenshot is provided, but the agent may express actions as a high-level triple \(a_t = (c, m, t)\), where \(c\) indexes the controlled character, \(m \in \{\text{Basic}, \text{Skill},
\text{Ultimate}, \text{Hold}\}\) specifies the move type, and \(t\) designates the target (e.g., an enemy identifier)[see Table \ref{tab:action_space}] A static UI map supplies fixed anchors for character icons and skill buttons. For targets that refer to on-screen entities, we attach a lightweight perception stack that runs on the current frame: a finetuned YOLOv8 detector (per task; \(\sim\)400 labeled battle images) returns, for each entity, a semantic label and a tight bounding box \((x,y,w,h)\) \cite{Terven_2023}, and PaddleOCR (zero-shot) extracts short texts such as HP\%, buff/debuff names, and remaining skill points \cite{li2022ppocrv3attemptsimprovementultra}. The detector outputs serve two roles simultaneously: (i) they “textify” the UI into compact structured tokens (enemy information, total damage dealt this turn, current HP of the four playable characters) that can be appended to the policy prompt to ease grounding; and (ii) they parameterize manipulation by providing the actionable region for \(t\). Action execution then maps \((c,m,t)\) to concrete OS events: the environment clicks the center of the detector’s bounding box for the referenced target. We will release the detection checkpoints and label schema for the structured TA observations to support reproducibility.

\begin{figure}[!t]
  \centering
  \includegraphics[width=\linewidth]{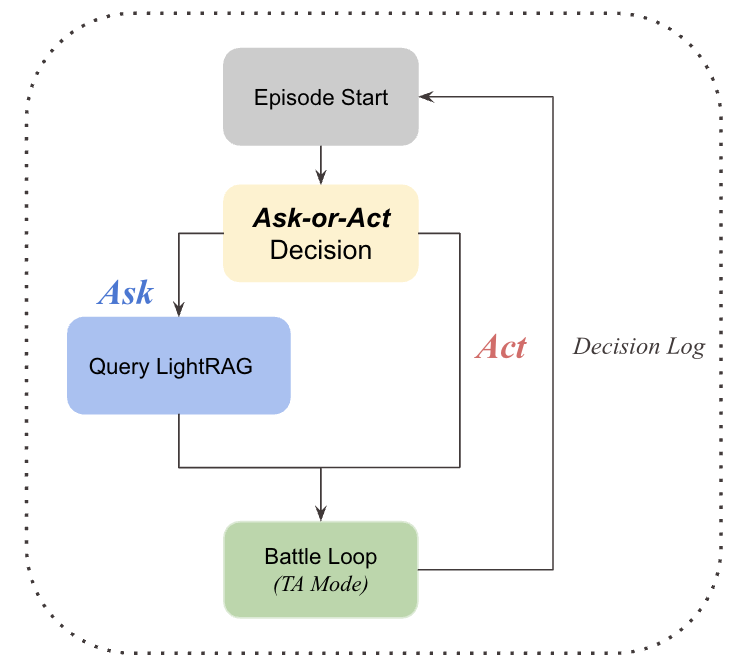}
  \caption{Ask--or--Act diagnostic pipeline. At episode start (TA regime), the agent chooses \textsc{Ask} (query LightRAG once for a short hint, persisted across the episode) or \textsc{Act} (proceed without guidance). Both paths run the same battle loop with the same action interface \(a_t=(c,m,t)\); they differ only in prompt context. A decision log records concise summaries (actions sequence for previous battle and battle result) to support learning-to-ask.}
  \label{fig:corpus_by_site}
\end{figure}


\subsection{Ask--or--Act Diagnostic}
\label{subsec:ask-or-act}
Humans seek help \emph{selectively}: they ask when expected benefit exceeds the (time/attention) cost and skip otherwise. Our diagnostic evaluates (i) \emph{efficiency}—how much performance is gained per unit cost of asking; and (ii) \emph{effect}—how much obtained guidance changes downstream performance (see \S\ref{subsec:metrics}).

\textit{StarBench} models the choice to seek guidance as a single pre-episode decision available \emph{only} in the tool-assisted (TA) regime, isolating decision making from low-level control. At the beginning of each battle, the agent chooses between \textsc{Act} (proceed immediately) or \textsc{Ask} (obtain one short textual hint, then proceed). We implement \textsc{Ask} via LightRAG \citep{guo2024lightrag}, instantiated with GPT-4o-mini \citep{openai2023gpt4}, over a fixed, read-only corpus of public HSR resources (wikis, guides, FAQs)[see Figure\ref{fig:corpus_by_site} for corpus composition].\footnote{The corpus is public-only and frozen for all runs. No coordinates, macros, or action strings are returned; responses are brief textual guidance (mechanics, priorities) only.} We will also release the LightRAG checkpoint for better reproduction of baseline results.

If \textsc{Ask} is selected, the agent issues a \emph{single}, targeted question before the episode starts. LightRAG returns one bounded textual answer, which is then \emph{persisted} and appended to the model prompt at \emph{every} subsequent decision step during the battle. No additional queries are permitted within the same episode. Execution is otherwise identical to TA (same screenshot, textualized observations, and action interface \(a_t=(c,m,t)\)). To support learning-to-ask, we also provide a decision log for each episode, enabling policies that calibrate the \textsc{Ask} threshold from past performance.

\begin{figure}[htbp]
  \centering
  \includegraphics[width=\linewidth]{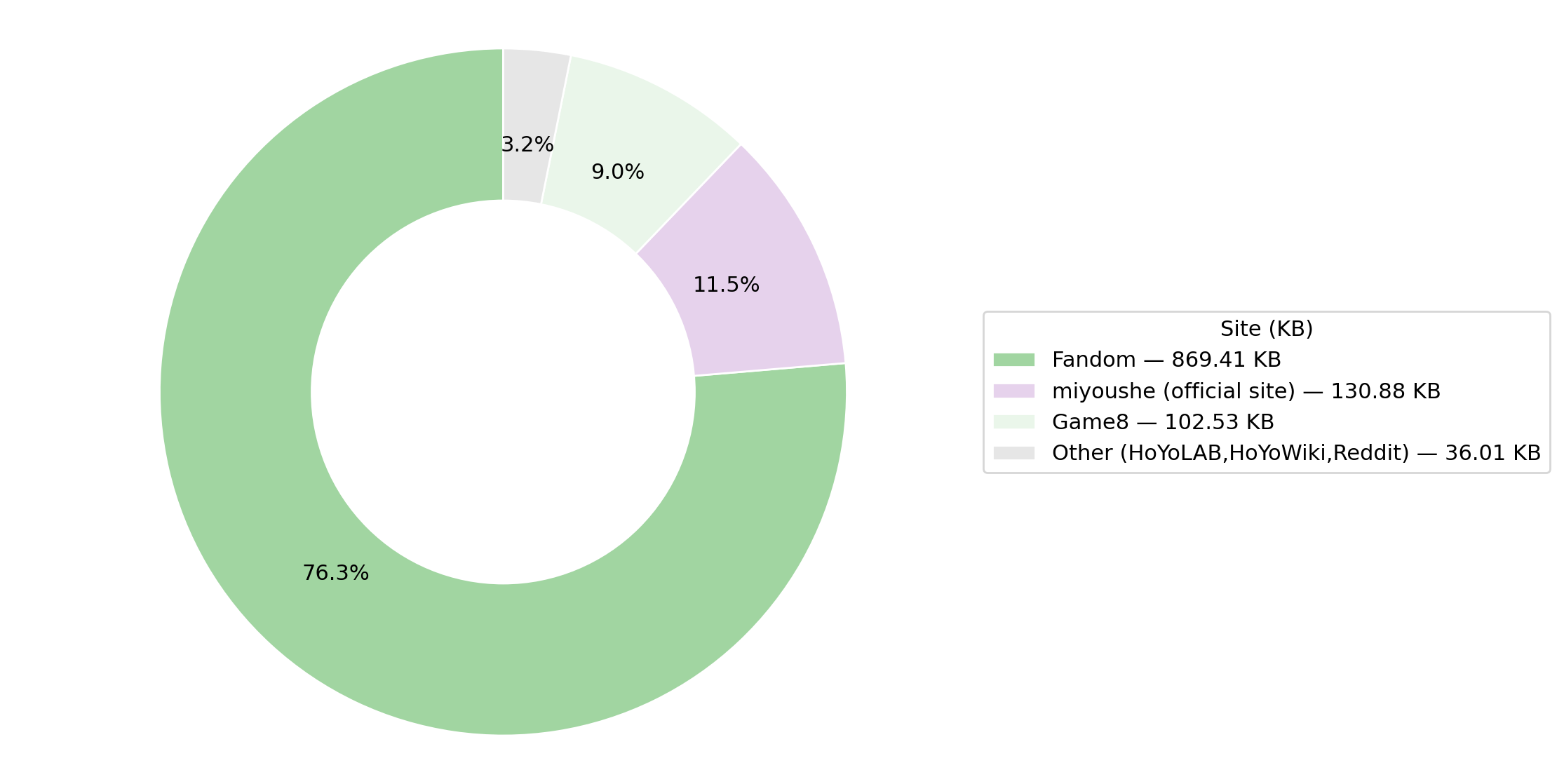}
  \caption{Corpus composition by \textbf{site}.
  \emph{Forums}: HoYoLAB\cite{hoyolab}, Reddit\cite{reddit};
  \emph{Wiki}: Fandom\cite{fandom}, Game8\cite{game8}, HoYoWiki\cite{hoyowiki};
  \emph{Official site}: Miyoushe\cite{miyoushe}.
  Total: 1.14\,MB (1138.39\,KB). 
  }
  \label{fig:corpus_by_site}
\end{figure}

\subsection{Task Suite}

\textit{StarBench} comprises eight battles that cover the four principal \emph{HSR} combat families (Table~\ref{tab:task_specification}). We adopt the game’s native objectives and report family-specific metrics as formalized in \S\ref{subsec:metrics}; here we summarize what each family tests, avoiding repetition of those definitions. \emph{Upper Stage} denotes activities with two stages (Upper/Lower) that require distinct teams; for reproducibility we standardize on the Upper Stage. \emph{No Assistance} follows the in-game rule (e.g., \emph{End of the Eternal Freeze} disables the Engine of Creation), so outcomes reflect the agent’s own control.

\subsubsection{Echo of War.}
Single-boss encounters that emphasize \emph{efficient finish}. Agents must parse the time skills/ultimates, and manage survivability while minimizing decision steps. 

\subsubsection{Memory of Chaos.}
Two-wave battles governed by limited time. The core challenge is \emph{tempo management}: planning damage and healing so both waves complete within a limited number of cycles.

\subsubsection{Pure Fiction.}
A score-chasing mode under a global time budget. Unlike the above, the focus is \emph{throughput}: rapidly clearing adds and chaining AOEs to maximize the in-game score within fixed time. 

\subsubsection{Apocalyptic Shadow.}
Boss fights scored by a composite of \emph{HP depletion} and \emph{remaining time}. This family stresses burst windows, mitigation, and end-of-fight economy.
\begin{table}[htbp]
  \centering
  \scriptsize
  \resizebox{\linewidth}{!}{
  \begin{tabular}{lcccc}
    \toprule
    \textbf{Task ID} & \textbf{Task Name} & \textbf{Family}  & \textbf{Notes} \\
    \midrule
    1 & cocolia       & Echo of War             & No Assistance \\
    2 & phantylia     & Echo of War                         & No Assistance \\
    3 & swarm king    & Echo of War          & -- \\
    4 & theater       & Echo of War           & No Assistance \\
    5 & feixiao       & Echo of War             & -- \\
    *6 & xianzhou     & Forgotten Hall                & Upper Stage \\
    *7 & cliched      & Pure Fiction                    & Upper Stage \\
    *8 & stardevourer & Apocalyptic Shadow           & Upper Stage \\
    \bottomrule
  \end{tabular}}
  \caption{Task specifications for \textit{StarBench}. Starred tasks include multi-wave or scoring-oriented constraints}
  \label{tab:task_specification}
\end{table}

\subsection{Evaluation Metrics}
\label{subsec:metrics}
In HSR, while many combat challenges only require characters to remain undefeated at the end, some combat challenges with high difficulties have additional requirements, including total “time” – in terms of the action values (AVs) – consumption and scores earned by defeating enemies. We adopt the game’s native scoring/timing for each activity and standardize only logging and aggregation. Preference is indicated with monotone arrows: $\downarrow$ (lower preferred), $\uparrow$ (higher preferred).

\begin{enumerate}[leftmargin=1.2em]
\item \textbf{Echo of War}. We report two metrics.

\noindent\emph{(i) Completion steps} (\(\downarrow\)):
\[
t_{\mathrm{steps}}\in\mathbb{N}\cup\{+\infty\},
\]
the number of agent decisions until victory. If the boss is not
defeated, set \(t_{\mathrm{steps}}=+\infty\).

\noindent\emph{(ii) Episode reward} (\(\uparrow\)):
\[
R_{\mathrm{EoW}}=\sum_{t=1}^{T} r_t,\qquad
r_t=\tfrac{1}{2}\,r^{\HP}_t+\tfrac{1}{2}\,r^{\DMG}_t,
\]
the total assigned reward by summing up the reward per action. 
For every action at $t$, the reward is assigned upon the HP condition of all characters ($r^{\HP}_{t}$) and the damage dealt by the active character ($r^{\DMG}_{t}$). For \(r^{\HP}_t\), rewards are assigned on effective and timely heal, particularly when characters are at risk of not surviving the next round of enemy attacks. We combine 
\begin{equation*}
\begin{aligned}
    \hat{\epsilon}^{\HP}_{t}\coloneqq \hat{\boldsymbol{h}}_{t}+\Delta \hat{\boldsymbol{h}}_{t-1},\quad
    r^{\text{safety}}_{t}&\coloneqq \big(\sign(\hat{\epsilon}^{\HP}_{t}) \geq \mathbf{0}\big), \\
    r^{\text{healing}}_{t}&\coloneqq \neg\big( \sign( \Delta \hat{\boldsymbol{h}}_{t} ) \geq \mathbf{0} \big).
\end{aligned}
\end{equation*}
into a boolean variable 
\begin{equation*}
    r^{h}_{t} \coloneqq r^{\text{safety}}_{t} + r^{\text{healing}}_{t} - \big(r^{\text{safety}}_{t} \cdot r^{\text{healing}}_{t}\big).
\end{equation*}
Using $r^{h}_{t}$, \(r^{\HP}_{t}\) is computed as
\begin{equation*}
\begin{array}{clcl}
    \hat{\delta}^{\HP}_{t} & \coloneqq \hat{\boldsymbol{h}}_{t} - \tfrac{1}{2}\,\mathbf{1}, &
    \hat{\sigma}^{\HP}_{t} & \coloneqq -\sign(\hat{\epsilon}^{\HP}_{t}) \odot \Delta \hat{\boldsymbol{h}}_{t},\\[0.25em]
    \hat{\Sigma}^{\HP}_{t} & \coloneqq (\neg r^{h}_{t}) \cdot \hat{\delta}^{\HP}_{t} + (r^{h}_{t})\cdot \hat{\sigma}^{\HP}_{t}, &
    r^{\HP}_{t} & = -\sign(\hat{\epsilon}^{\HP}_{t}) \cdot \hat{\Sigma}^{\HP}_{t}.
\end{array}
\end{equation*}
The damage term is
\[
r^{\DMG}_t \;=\; \hat d_t.
\]

Notation: \(\hat{\boldsymbol{h}}_{t}\): normalized HP vector at step \(t\), with all characters concatenated;
\(\Delta \hat{\boldsymbol{h}}_{t}\): change in \(\boldsymbol{h}\) from the previous timestep; 
\(\hat d_t\): normalized damage at \(t\); $\mathbf{0}$ and $\mathbf{1}$ are zero and one vectors, respectively.

\item \textbf{Memory of Chaos}. Cycles advance by the game’s AV schedule (first cycle: \(150\) AV; subsequent: \(100\) AV). With total AV consumed \(\mathrm{AV}_{\mathrm{used}}\),
\begin{equation}
    C_{\mathrm{used}}
    =
    \mathbf{1}\{\mathrm{AV}_{\mathrm{used}} \ge 150\}
    +
    \max\!\left(0,\;\left\lfloor \frac{\mathrm{AV}_{\mathrm{used}}-150}{100} \right\rfloor\right),
\end{equation}
and with cycle budget \(C_{\max}\), we report remaining cycles
\begin{equation}
    S_{\mathrm{MoC}}
    \;=\;
    \max\!\big(0,\; C_{\max} - C_{\mathrm{used}} \big)
    \qquad (\uparrow).
\end{equation}

\item \textbf{Pure Fiction}. With global AV budget \(\mathrm{AV}_{\max}=450\) and in-game incremental score \(s_t\) from enemy eliminations at step \(t\),
\begin{equation}
    S_{\mathrm{PF}}
    \;=\;
    \sum_{t=1}^{T} s_t \cdot \mathbf{1}\{\mathrm{AV}_t \le \mathrm{AV}_{\max}\}
    \qquad (\uparrow).
\end{equation}

\item \textbf{Apocalyptic Shadow}. Let \(\mathcal{S}_{\mathrm{AS}}(\mathrm{HP\%}_{\downarrow}, \mathrm{AV}_{\mathrm{rem}})\) be the game’s native composite score (boss HP depletion + remaining AV bonus). We report
\begin{equation}
    S_{\mathrm{AS}}
    \;=\;
    \mathcal{S}_{\mathrm{AS}}(\mathrm{HP\%}_{\downarrow}, \mathrm{AV}_{\mathrm{rem}})
    \qquad (\uparrow).
\end{equation}

\item \textbf{Ask--or--Act Metrics}.
\label{subsubsec:ask-metrics}
Let \(S\) be the task score; for \emph{Echo of War} use \(S=-\mathrm{Steps}\).
Episodes are ordered \emph{within each task} by time: \(k=1,\dots,n_t\).
Let \(A_{t,k}\in\{0,1\}\) indicate \textsc{Act} (0) vs.\ \textsc{Ask} (1).
Define the ask rate \(\displaystyle \mathrm{AR}=\frac{1}{N}\sum_{t,k} A_{t,k}\), where \(N\) is the number of evaluated episodes, and
let \(M=\sum_{t,k}\mathbf{1}\{A_{t,k}=1,\,k>1\}\) be the count of asked episodes that have a predecessor.

\noindent\emph{(i) Effect (temporal uplift per ask; \(\uparrow\)).}
For every asked episode with a predecessor on the same task (\(A_{t,k}=1\) and \(k>1\)), set
\[
\Delta_{t,k} \;=\; S_{t,k} - S_{t,k-1}.
\]
Report the mean per-ask effect:
\[
\widehat{\mathrm{Effect}} \;=\; \frac{1}{M}\sum_{t}\sum_{k:\,A_{t,k}=1,\,k>1} \Delta_{t,k}.
\]

\noindent\emph{(ii) Efficiency (normalized by number of asks; \(\uparrow\)).}
Each evaluation allows at most \(T\) asks. The expected number of asks is \((T\cdot\mathrm{AR}/100)\).
We therefore define
\[
\mathrm{Efficiency} \;=\; \frac{\widehat{\mathrm{Effect}}}{(T\cdot \mathrm{AR}/100)}
\,
\]

\end{enumerate}

\newcommand{\bestup}{\textsuperscript{$\blacktriangle$}}      
\newcommand{\secondup}{\textsuperscript{$\triangle$}}         
\newcommand{\bestdown}{\textsuperscript{$\blacktriangledown$}}
\newcommand{\seconddown}{\textsuperscript{$\triangledown$}}   

\begin{table*}[t]
\centering
\small
\setlength{\tabcolsep}{4.5pt}
\caption{\textbf{Echo of War per-task results.}
We report \textbf{SR\%} (\(\uparrow\)), \textbf{Steps} (\(\downarrow\)), and \(\boldsymbol{R_{\mathrm{EoW}}}\) (\(\uparrow\)); rewards use a unified 0--100 scale. VLMs are grouped by regime; baselines appear separately. \textbf{Coloring:} best VLM per boss in \textit{yellow}, best Baseline per boss in \textit{blue}; the winner’s SR/Steps/Reward cells are all highlighted. \textbf{Tie-break:} higher SR \(\rightarrow\) lower Steps \(\rightarrow\) higher Reward. \(\infty\) stands for infinite steps for failure case and would not take in count during comparison.} 
\label{tab:eow-perboss-3metrics}
\begin{adjustbox}{max width=\linewidth}
\begin{tabular}{l ccc ccc ccc ccc ccc}
\toprule
& \multicolumn{3}{c}{\textbf{Cocolia}} & \multicolumn{3}{c}{\textbf{Phantylia}} & \multicolumn{3}{c}{\textbf{Swarm King}}
& \multicolumn{3}{c}{\textbf{Theater}} & \multicolumn{3}{c}{\textbf{Feixiao}} \\
\cmidrule(lr){2-4}\cmidrule(lr){5-7}\cmidrule(lr){8-10}\cmidrule(lr){11-13}\cmidrule(lr){14-16}
\textbf{Model} &
\(\text{SR}\%\) & \(\text{Steps}\) & \(R_{\mathrm{EoW}}\) &
\(\text{SR}\%\) & \(\text{Steps}\) & \(R_{\mathrm{EoW}}\) &
\(\text{SR}\%\) & \(\text{Steps}\) & \(R_{\mathrm{EoW}}\) &
\(\text{SR}\%\) & \(\text{Steps}\) &\(R_{\mathrm{EoW}}\) &
\(\text{SR}\%\) & \(\text{Steps}\) & \(R_{\mathrm{EoW}}\) \\
\midrule
\multicolumn{16}{l}{\textit{VLMs — Direct Control (DC: screenshot$\rightarrow$OS primitives, no text aids)}}\\
GPT\textendash4o-mini (DC) &
0.0 & \(\infty\) & \(0.14\!\pm\!0.0\) &
0.0 & \(\infty\) & \(0.06\!\pm\!0.0\) &
0.0 & \(\infty\) & \(0.12\!\pm\!0.0\) &
0.0 & \(\infty\) & \(0.10\!\pm\!0.0\) &
0.0 & \(\infty\) & \(0.11\!\pm\!0.0\) \\
Claude 3.5 Sonnet (DC) &
0.0 & \(\infty\) & \(0.08\!\pm\!0.0\) &
0.0 & \(\infty\) & \(0.07\!\pm\!0.0\) &
0.0 & \(\infty\) & \(0.11\!\pm\!0.0\) &
0.0 & \(\infty\) & \(0.09\!\pm\!0.0\) &
0.0 & \(\infty\) & \(0.10\!\pm\!0.0\) \\
Gemini 1.5 Flash (DC) &
0.0 & \(\infty\) & \(0.0\!\pm\!0.0\) &
0.0 & \(\infty\) & \(0.0\!\pm\!0.0\) &
0.0 & \(\infty\) & \(0.0\!\pm\!0.0\) &
0.0 & \(\infty\) & \(0.0\!\pm\!0.0\) &
0.0 & \(\infty\) & \(0.0\!\pm\!0.0\) \\
\midrule
\multicolumn{16}{l}{\textit{VLMs — Tool-Assisted (TA: \(a_t=(c,m,t)\) with YOLO/ optional OCR text helps)}}\\
GPT\textendash4o-mini (TA\textendash NO\textendash OCR) &
62.5 & \(44.3\!\pm\!6.8\) & \(63.2\!\pm\!4.1\) &
37.5 & \(92.0\!\pm\!12.3\) & \(50.8\!\pm\!3.7\) &
12.5 & \(128.0\!\pm\!18.5\) & \(46.3\!\pm\!4.9\) &
25.0 & \(132.7\!\pm\!15.4\) & \(45.1\!\pm\!3.2\) &
25.0 & \(141.5\!\pm\!20.6\) & \(44.2\!\pm\!3.1\) \\
Claude 3.5 Sonnet (TA\textendash NO\textendash OCR) &
62.5 & \(43.8\!\pm\!7.2\) & \(62.7\!\pm\!4.9\) &
25.0 & \(98.4\!\pm\!13.6\) & \(49.6\!\pm\!3.5\) &
12.5 & \(134.2\!\pm\!17.1\) & \(45.5\!\pm\!3.8\) &
37.5 & \(124.9\!\pm\!14.0\) & \(46.2\!\pm\!2.9\) &
25.0 & \(138.7\!\pm\!18.3\) & \(45.0\!\pm\!3.0\) \\

Gemini 1.5 Flash (TA\textendash NO\textendash OCR) &
12.5  & \(36.7\!\pm\!0.0\)   & \(67.1\!\pm\!0.0\) &
0  & \(\infty\)   & \(0.13\!\pm\!0.02\) &
0.0   & \(\infty\)           & \(0.0\!\pm\!0.4\) &
0.0   & \(\infty\)           & \(0.0\!\pm\!0.2\) &
0  & \(\infty\) & \(0.1\!\pm\!0.3\) \\
GPT\textendash4o-mini (TA) &
100.0 & \(31.40\!\pm\!2.60\) & \(71.1\!\pm\!1.8\) &
100.0 & \(50.8\!\pm\!6.54\) & \(59.7\!\pm\!2.6\) &
50.0  & \(70.8\!\pm\!3.20\) & \(54.1\!\pm\!5.1\) &
100.0 & \(84.6\!\pm\!11.02\) & \(52.0\!\pm\!1.3\) &
\wvlm{100.0} & \wvlm{\(83.1\!\pm\!7.26\)} & \wvlm{\(52.1\!\pm\!1.0\)} \\
Claude 3.5 Sonnet (TA) &
100.0 & \(31.50\!\pm\!2.00\) & \(71.5\!\pm\!5.6\) &
87.5  & \(47.0\!\pm\!6.0\)   & \(56.5\!\pm\!3.0\) &
37.5  & \(69.3\!\pm\!4.70\) & \(52.4\!\pm\!3.9\) &
87.5  & \(74.0\!\pm\!8.0\)   & \(51.0\!\pm\!3.0\) &
62.5  & \(100.0\!\pm\!10.0\) & \(54.0\!\pm\!3.0\) \\
Gemini 1.5 Flash (TA) &
12.5  & \(37.0\!\pm\!0.0\)   & \(67.1\!\pm\!0.0\) &
12.5  & \(54.0\!\pm\!9.0\)   & \(58.0\!\pm\!2.0\) &
0.0   & \(\infty\)           & \(-0.2\!\pm\!0.4\) &
0.0   & \(\infty\)           & \(3.0\!\pm\!1.0\) &
12.5  & \(132.0\!\pm\!10.0\) & \(47.0\!\pm\!2.0\) \\
GPT\textendash4o-mini (ReAct--TA) &
75.0  & \(30.20\!\pm\!1.20\) & \(69.8\!\pm\!5.6\) &
50.0  & \(51.8\!\pm\!5.60\) & \(56.8\!\pm\!4.5\) &
25.0  & \(70.5\!\pm\!3.50\) & \(50.6\!\pm\!2.9\) &
62.5  & \(76.3\!\pm\!3.90\) & \(53.1\!\pm\!0.7\) &
75.0  & \(76.40\!\pm\!7.50\) & \(53.3\!\pm\!1.2\) \\
GPT\textendash4o-mini (Reflexion--TA) &
62.5 & \(30.4\!\pm\!3.4\) & \(68.4\!\pm\!7.2\) &
50.0 & \(52.4\!\pm\!7.8\) & \(54.3\!\pm\!3.5\) &
\wvlm{62.5} & \wvlm{\(72\!\pm\!10.2\)} & \wvlm{\(49.2\!\pm\!2.8\)} &
62.5 & \(75.3\!\pm\!4.25\) & \(52.8\!\pm\!1.2\) &
62.5 & \(84.4\!\pm\!11.5\) & \(52.4\!\pm\!2.4\) \\
GPT\textendash4o-mini (TA-Ask) &
87.5 & \(29.4\!\pm\!4.2\) & \(69.2\!\pm\!5.7\) &
62.5 & \(55.0\!\pm\!8.3\) & \(58.2\!\pm\!2.3\) &
50.0 & \(69.3\!\pm\!9.8\) & \(54.5\!\pm\!2.1\) &
100.0 & \(74.1\!\pm\!9.0\) & \(51.4\!\pm\!4.8\) &
100.0 & \(119.4\!\pm\!10.3\) & \(47.1\!\pm\!4.7\) \\
Claude 3.5 Sonnet (TA-Ask) &
\wvlm{100.0} & \wvlm{\(30.6\!\pm\!4.8\)} & \wvlm{\(69.1\!\pm\!8.0\)} &
\wvlm{100.0} & \wvlm{\(44.1\!\pm\!3.8\)} & \wvlm{\(58.5\!\pm\!6.0\)} &
50.0 & \(65.7\!\pm\!10.3\) & \(55.6\!\pm\!2.0\) &
\wvlm{100.0} & \wvlm{\(70.8\!\pm\!9.7\)} & \wvlm{\(52.9\!\pm\!3.7\)} &
75.0 & \(96.5\!\pm\!14.0\) & \(55.9\!\pm\!2.0\) \\
Gemini 1.5 Flash (TA-Ask) &
12.5 & \(39.0\!\pm\!0.0\) & \(65.5\!\pm\!0.0\) &
25.0 & \(51.5\!\pm\!9.2\) & \(59.3\!\pm\!3.4\) &
0.0 & \(\infty\) & \(0.1\!\pm\!0.0\) &
0.0 & \(\infty\) & \(4.1\!\pm\!0.0\) &
12.5 & \(127.0\!\pm\!0.0\) & \(48.2\!\pm\!0.0\) \\
\midrule
\multicolumn{16}{l}{\textit{Baselines (no DC/TA separation; Human participates, reward missing)}}\\
Random (TA) &
25.0 & \(45.3\!\pm\!11.3\) & \(60.5\!\pm\!9.7\) &
0.0 & \(\infty\) & \(-0.4\!\pm\!0.3\) &
0.0 & \(\infty\) & \(0.0\!\pm\!0.1\) &
0.0 & \(\infty\) & \(-0.4\!\pm\!0.4\) &
0.0 & \(\infty\) & \(-0.2\!\pm\!0.2\) \\
Human (reference) &
\wbase{100.0} & \wbase{\(30.3\!\pm\!6.6\)} & \wbase{--} &
\wbase{90.0} & \wbase{\(43.8\!\pm\!11.6\)} & \wbase{--} &
\wbase{90.0} & \wbase{\(52.4\!\pm\!12.4\)} & \wbase{--} &
\wbase{90.0} & \wbase{\(77.0\!\pm\!14.4\)} & \wbase{--} &
\wbase{100.0} & \wbase{\(75.9\!\pm\!9.2\)} & \wbase{--} \\
\bottomrule
\end{tabular}
\end{adjustbox}
\end{table*}

\begin{table}[t]
\centering
\caption{\textbf{MoC / PF / AS family metrics.} Models shown with DC, TA, and TA-Ask subrows; baselines listed separately. -\(\infty\) stands for model failed to accomplished the MoC task} 
\label{tab:other-families-compact}
\scriptsize
\resizebox{\linewidth}{!}{
\begin{tabular}{l l rrr}
\toprule
\textbf{Model} & \textbf{Regime} & \textbf{MoC} $C_{\mathrm{rem}}\!\uparrow$ & \textbf{PF} Score $\uparrow$ & \textbf{AS} Score $\uparrow$ \\
\midrule
\multirow{3}{*}{GPT-4o-mini}
  & \reg{DC}     & -\(\infty\) & 0 & 0 \\
  & \reg{TA}     & $16\pm0.5$ & $7246\pm128$ & $2804\pm137$ \\
  & \reg{TA-Ask} & \wvlm{$18\pm0.5$} &  \wvlm{$10280\pm1348$} & $2810\pm150$ \\
\midrule
\multirow{3}{*}{Claude 3.5 Sonnet}
  & \reg{DC}     & -\(\infty\) & 0 & 0 \\
  & \reg{TA}     & $16\pm0.5$ & $7235\pm837$ & \wvlm{$2833\pm168$} \\
  & \reg{TA-Ask} & $16\pm0.5$ & $8000\pm246$ & $2825\pm125$ \\
\midrule
\multirow{3}{*}{Gemini 1.5 Flash}
  & \reg{DC}     & -\(\infty\) & 0 & 0 \\
  & \reg{TA}     & -\(\infty\) & $1780\pm374$ & $476\pm236$ \\
  & \reg{TA-Ask} & -\(\infty\) & $2000\pm120$ & $380\pm202$ \\
\midrule
Human (reference) & -- & $17\pm1.5$ & \wbase{$7560\pm1749$} & \wbase{$3240\pm170$} \\
Auto-battle       & -- & \wbase{$18\pm0.0$} & $7120\pm0.0$  & $3202\pm0.0$  \\
\bottomrule
\end{tabular}
}
\end{table}

\begin{table}[t]
\centering
\caption{\textbf{Ask--or--Act diagnostic (TA only).} \textbf{Effect} is the mean per-ask uplift vs.\ the immediately previous episode on the same task; \textbf{Efficiency} uses the normalization \(\mathrm{Effect}/[(\mathrm{AR}/100)\times 8]\). AR shown as percent. Best effect in yellow, while best efficiency in blue.}
\label{tab:ask-or-act}
\resizebox{\linewidth}{!}{
\begin{tabular}{l l r r r}
\toprule
\textbf{Model} & \textbf{Task} & \textbf{AR (\%)} & \textbf{Effect} & \textbf{Efficiency} \\
\midrule
\cmidrule(lr){1-5}
GPT\textendash4o-mini             & Echo of War          & 52.5 & \wvlm{82.8}  & 19.7 \\
Claude 3.5 Sonnet         & Echo of War          & 22.5 & 49.4  & \wbase{27.4} \\
Gemini 1.5 Flash                  & Echo of War          & 97.5 & 47.4  & 6.1  \\
\midrule
\cmidrule(lr){1-5}
GPT\textendash4o-mini     & Memory of Chaos      & 62.5 & \wvlm{17}    & \wbase{3.4}  \\
Claude 3.5 Sonnet                 & Memory of Chaos      & 100  & 16    & 2.0  \\
Gemini 1.5 Flash                  & Memory of Chaos      & 100  & 0     & 0.0  \\
\midrule
\cmidrule(lr){1-5}
GPT\textendash4o-mini      & Pure Fiction         & 100  & \wvlm{10800} & \wbase{1350.0} \\
Claude 3.5 Sonnet                 & Pure Fiction         & 100  & 6400  & 800.0  \\
Gemini 1.5 Flash                  & Pure Fiction         & 100  & 2000  & 250.0  \\
\midrule
\cmidrule(lr){1-5}
GPT\textendash4o-mini      & Apocalyptic Shadow   & 100  & \wvlm{3002}  & \wbase{375.3} \\
Claude 3.5 Sonnet                 & Apocalyptic Shadow   & 100  & 2817  & 352.1 \\
Gemini 1.5 Flash                  & Apocalyptic Shadow   & 100  & 380   & 47.5  \\
\bottomrule
\end{tabular}}
\raggedright\footnotesize
\(T=8\) ask opportunities per evaluation. Efficiency is set to 0 when \(\mathrm{AR}=0\).
\end{table}

\section{Experiment}
\subsection{Baselines} 
\label{eval:baseline}
We build several baselines in StarBench: a random policy, human expert performance, AutoBattle , a multimodal RL policy with VLMs baselines (GPT4O-mini,Claude-3.5-Sonnet, Gemini-1.5-Flash,). with two prompt engineering method: ReAct and Reflexion
\subsubsection{Random Policy} This policy selects a valid action uniformly at random from the set of permissible actions indicated by the action mask in the observation with tool-assisted mode. It does not exploit any visual or textual cues.
\subsubsection{Human Expert} We recruited 10 participants with different levels of experience in HSR gameplay, ranging from novices with no previous exposure to veterans with over 200 hours of gameplay, to engage in controlled battle sessions. Their performance was recorded and averaged to establish a human performance benchmark.
\subsubsection{AutoBattle}
AutoBattle is a function in \emph{Honkai: Star Rail} that automatically controls the player's team during combat. The usage and targeting of abilities are determined by many factors, such as skill points, enemy weakness, enemy HP, status effects, etc. AutoBattle is more like a state machine that follows simple logic. For example, it will release the Ultimate right when it is ready, except such Ultimates with healing behaviors \citep{honkaiAutoBattle}.

\subsubsection{Prompting Variants: ReAct and Reflexion}
In addition to evaluating the VLMs directly, we also assess two lightweight decoding-time prompting strategies:\emph{ReAct} (Reasoning+Acting) \cite{yao2023react} and \emph{Reflexion} (Self-critique) \cite{shinn2023reflexion}.

\subsection{Experiment setup}
For each evaluation, we run each agent \textbf{eight} trials per task using the same standardized team configuration. All experiments execute on a single RTX~4070~Ti GPU. The HSR client runs windowed at 1920$\times$1080 with UI scale 100\%, VSync on, and a fixed graphics preset. Input events are issued via OS-level primitives with a fixed 0.5\,s inter-event delay.

Each task emphasizes turn-by-turn decision complexity, including optimal skill-point usage and effective exploitation of enemy weakness breaks. An episode terminates on victory, failure, or invalid-output conditions. For \textbf{DC}, if the agent fails to produce a valid OS primitives on a required UI widget for $10$ consecutive attempts (e.g., off-target region, or empty emission), we stop the episode and record a failure. For \textbf{TA}, if the agent emits an invalid tuple $(c,m,t)$ that cannot be mapped to a legal action or violates preconditions (e.g., attempting \textsc{Skill} with zero skill points) for $K{=}10$ consecutive decisions, we also stop and record a failure. Timeout decisions are treated as no-ops and count toward step/AV budgets.

\subsection{Results}
StarBench evaluates two human-like competencies introduced in the abstract and introduction: (i) pixel-to-action grounding—mapping raw screenshots to precise OS primitives; and (ii) agentic information seeking—deciding when brief guidance is worth the cost. By holding tasks and metrics fixed across regimes, Tables~\ref{tab:eow-perboss-3metrics}, \ref{tab:other-families-compact}, and \ref{tab:ask-or-act} isolate where capability breaks.

\textbf{(1) DC exposes a fundamental grounding gap.}
In Table~\ref{tab:eow-perboss-3metrics} (DC blocks), GPT\textendash4o-mini, Claude~3.5~Sonnet, and Gemini~1.5~Flash all record \(\text{SR}=0\%\) and near-zero \(R_{\mathrm{EoW}}\) on every Echo of War boss; Table~\ref{tab:other-families-compact} shows the same collapse for MoC (\(-\infty\) cycles), PF (0), and AS (0). With tasks/metrics unchanged, these failures specifically indicate an end-to-end screenshot\(\rightarrow\)primitive localization deficit rather than task difficulty. For DC, most of cases failed because those VLMs selected incorrect space for targets (e.g., selecting the non-UI elements in the battle scene). As results, most of actions are selected by VLMs under DC mode are choosing the default target and finally lead to bad performance (e.g., keep using the basic attacks to attack left most enemies and never use skill).

\textbf{(2) Minimal UI grounding unlocks competent play under the same tasks.}
Switching only the interface to TA yields large gains. In Table~\ref{tab:eow-perboss-3metrics} (TA blocks), GPT\textendash4o-mini reaches \(100\%\) SR on \emph{Cocolia}, \emph{Phantylia}, \emph{Theater}, and \emph{Feixiao} with finite steps on \emph{Cocolia}, on \emph{Phantylia}), and Claude shows a similar pattern. These improvements carry to other families in Table~\ref{tab:other-families-compact}. The key reason for this performance jump is the abstraction provided by the TA regime: instead of issuing low-level OS primitives (e.g., pixel-level clicks), VLMs in TA mode operate over structured action tuples (character, move type, target). The model can focus on higher-level decision-making—such as when to cast a Skill or Ultimate—without worrying about where to click or whether the UI has changed slightly. For example, under TA, models reliably trigger the correct abilities because the action space is cleanly defined. In contrast, in the DC regime, even if a model correctly reasons that an Ultimate should be used, it may fail to locate and click the correct UI element—resulting in a no-op or illegal action. Thus, TA removes a major bottleneck: the challenge of grounding visual perception to actionable control, allowing evaluation to better isolate strategic competence from UI manipulation skill.

\textbf{(3) Why OCR matters beyond boxes.}
Removing OCR while keeping YOLO bounding boxes consistently hurts both success and efficiency (Table~\ref{tab:eow-perboss-3metrics}, TA\textendash NO\textendash OCR). For GPT\textendash4o-mini, SR drops from \(100\%\) to \(62.5\%\) on \emph{Cocolia}, to \(37.5\%\) on \emph{Phantylia}, to \(25.0\%\) on \emph{Theater} and \emph{Feixiao}; Steps inflate (e.g., \emph{Cocolia} \(31.4\!\rightarrow\!44.3\), \emph{Phantylia} \(50.8\!\rightarrow\!92.0\)) and \(R_{\mathrm{EoW}}\) falls accordingly. Claude shows parallel degradations (e.g., \emph{Cocolia} \(100\%\!\rightarrow\!62.5\%\), \emph{Phantylia} \(87.5\%\!\rightarrow\!25.0\%\)). These gaps indicate that textualized UI (HP\%, skill points, readiness/status strings) reduces decision friction—especially for legality and timing—beyond what spatial boxes provide. Practically, this also surfaces a limitation of native VLM text recognition on dense, stylized UIs: without OCR, models misread or miss small texts(e.g., characters' HP), leading to illegal or mistimed actions and higher Steps (e.g., forgot to use Healing Ultimate when characters' HP are low).

\textbf{(4) Asking is beneficial when calibrated.}
Table~\ref{tab:ask-or-act} shows that brief guidance produces measurable uplifts when used judiciously. GPT\textendash4o-mini attains the largest \emph{Effect} across Echo of War (\(82.8\)), PF (\(10800\)), and AS (\(3002\)), while Claude achieves the best \emph{Efficiency} on Echo of War (\(27.4\)) despite a lower ask rate. Gemini asks frequently (AR \(97.5\%\)) but gains little (Effect \(47.4\), Efficiency \(6.1\)). We observe that all VLMs tend to ask questions during the first trial of a task (out of eight total), likely due to uncertainty about unfamiliar mechanics or objectives. For tasks with explicit goals—such as Echo of War (defeating a boss)—models typically delay asking until after a failure, suggesting a reactive asking policy. However, for implicit-goal tasks, such as maximizing scores in Pure Fiction or managing time in Memory of Chaos, models ask more proactively and frequently, likely in an attempt to iteratively refine strategy across episodes.

\section{Conclusion}
We introduced \textbf{StarBench}, a real-client benchmark for testing whether vision--language models can (i) ground raw pixels into coherent keyboard–mouse actions and (ii) decide when to seek guidance via an \emph{ask-or-act} diagnostic. By fixing tasks and metrics and varying only the interaction regime—\emph{direct} screenshot-to-action versus \emph{tool-assisted} control—StarBench cleanly separates perception-to-control grounding from higher-level decision making. Current VLMs fail almost entirely in direct control, revealing a core deficiency in pixel-to-primitive localization and UI manipulation. Tool assistance, especially with textualized UI, markedly improves success beyond that. The ask-or-act results further demonstrate that brief, targeted guidance can yield measurable uplifts. These findings support that competent real-client play today hinges on lightweight tools and calibrated information seeking, while end-to- end pixel-to-primitive control remains the key challenge.

\section{Limitations and Future Work}
StarBench runs on a \emph{real} live-service client, which introduces practical constraints (frame-timing variance, and OS-level input flakiness) that may affect exact reproducibility. To mitigate these issues, we are developing a companion \emph{emulator} that replays recorded traces. This will allow frozen UI states, deterministic screenshots, and offline evaluation while keeping the same tasks and metrics.





\bibliographystyle{ACM-Reference-Format} 
\bibliography{sample}


\end{document}